\documentclass{article}
\usepackage{microtype}
\usepackage{graphicx}
\usepackage{subfigure}
\usepackage{booktabs} 

\usepackage{hyperref}

\usepackage[accepted]{icml2025}

\usepackage{amsmath}
\usepackage{amssymb}
\usepackage{mathtools}
\usepackage{amsthm}

\usepackage[capitalize,noabbrev]{cleveref}

\theoremstyle{plain}

\theoremstyle{definition}

\theoremstyle{remark}

\usepackage[textsize=tiny]{todonotes}

\makeatletter
\renewcommand{\@pa}[1]{}                                
\renewcommand{\printAffiliationsAndNotice}[1]{}         
\renewcommand{\icmlauthor}[2]{\par{\bf #1}\\#2\par}     
\makeatother

\begin{document}

\twocolumn[
\icmltitle{Leveraging Manifold Embeddings for Enhanced Graph Transformer Representations and Learning}




\begin{icmlauthorlist}
\icmlauthor{\textbf{Ankit Jyothish} \textnormal{and} \textbf{Ali Jannesari}}{}
\end{icmlauthorlist}

\centerline{Iowa State University, Ames, IA, USA}
\centerline{\texttt{\{ankitj99, jannesari\}@iastate.edu}}



\icmlkeywords{Machine Learning, ICML}

\vskip 0.3in
]



\printAffiliationsAndNotice{\icmlEqualContribution} 

\begin{abstract}
Graph transformers typically embed every node in a single Euclidean space, blurring heterogeneous topologies. We prepend a lightweight Riemannian mixture-of-experts layer that routes each node to various kinds of manifold, mixture of spherical, flat, hyperbolic—best matching its local structure. These projections provide intrinsic geometric explanations to the latent space. Inserted into a state-of-the-art ensemble graph transformer, this projector lifts accuracy by up to 3\% on four node-classification benchmarks. The ensemble makes sure that both euclidean and non-euclidean features are captured. Explicit, geometry-aware projection thus sharpens predictive power while making graph representations more interpretable.

\end{abstract}

\section{Introduction}
Graph‐structured data pervade domains such as molecular design, recommender systems, and social or mobile networks, where relational information drives tasks like node classification, link prediction, and substructure discovery. Progress hinges on pairing a high-capacity learning model with node and edge representations that faithfully encode both local context and global topology, thereby yielding accurate and interpretable predictions. Yet a uniform embedding space rarely suffices: heterogeneous graphs exhibit widely varying curvatures and connective motifs, calling for representations that adapt to each region’s geometry.

We address this gap by uniting fine-grained, node-specific Riemannian embeddings with a strong ensemble-based graph Transformer. The manifold component tailors curvature to local structure, while the Transformer provides scalable global reasoning \& GNN captures local euclidean features—together capturing subtle topological patterns that elude conventional methods. Extensive experiments on benchmark node-classification datasets confirm the effectiveness of this synergy. Our main contributions are:
\begin{itemize}
\item \textbf{Novel architecture},: First to combine granular Riemannian node embeddings with an ensemble-style graph Transformer backbone.
\item \textbf{Adaptive expressivity},: Node-level curvature selection enhances structural fidelity, enabling the model to learn from diverse graph regions without manual tuning.
\item \textbf{Empirical gains},: Achieves up to a 3\% accuracy improvement over strong baselines on standard node-classification benchmarks.
\end{itemize}

\section{Related Works}
\noindent Graph neural network (GNN) research began with spectral and spatial convolutions such as the Graph Convolutional Network (GCN)~\cite{gcn}, which aggregate information from a node’s immediate neighborhood.  
Successive refinements introduced neighborhood sampling~\cite{sage,neighborhoodsampling}, graph partitioning~\cite{graphpartition}, and knowledge–distillation or historical-embedding tricks~\cite{graphless,historicalembeddings} to curb computational blow-up.  
Attention-based Graph Attention Networks (GAT)~\cite{gat} then dominated, dynamically re-weighting neighbours and inspiring fully attention-driven variants~\cite{onlyattention}.  
Yet stacking many message-passing layers still multiplies cost with depth.  
Remedies include bi-kernel gating that handles both homophilic and heterophilic graphs~\cite{gatedbikernel} and pipelines where MLPs train fast while GNNs infer efficiently~\cite{mlpgraph}.

\noindent Graph Transformers push beyond local propagation by treating the graph as a fully connected token sequence and learning long-range dependencies via global self-attention~\cite{gt1,gtbert}.  
The $\mathcal{O}(N^{2})$ memory and runtime of this mechanism, however, limit adoption on million-node graphs.  
Efficiency-oriented variants therefore rely on node or ego-graph sampling, kernelized Gumbel-Softmax~\cite{nodeformer}, diffusion-based linear attention~\cite{difformer}, or spectral positional encodings~\cite{spectralattention}.  
Although near-linear in complexity, these approximations may lose expressivity or fail at web-scale.  
SGFormer~\cite{sgformer} resolves the tension by coupling a lightweight global linear-attention module with a standard GNN for local detail, and we adopt it as the backbone of our architecture.

\section{Limitations of Embedding Strategies}
Most graph‐representation models still map nodes to flat Euclidean space, even though real graphs often exhibit deep hierarchies and curved structure. Early fixes—such as spectral–spatial hybrids and algorithms like PageRank that probe latent topology \cite{pageranktopology,harmonicgraph}—highlighted the need for non-Euclidean geometry. Riemannian embeddings answered that call, yet constant-curvature settings \cite{constantcurvature,hgcn} prove too rigid for heterogeneous graphs. Product manifolds, Q-GCN, and Lorentzian convolutions \cite{qgcn,lorentz} ease this constraint, while motif-aware encoders add local curvature diversity \cite{motifrgc}. Mixture-of-experts schemes further enrich expressiveness by gating across hops or experts \cite{graphmixtureofexperts}, and have even been used to diagnose out-of-distribution shifts \cite{graphMETRO}.

The newest wave focuses on tailoring geometry to each node. GraphMoRE \cite{graphmore} assigns personalized manifolds that seamlessly blend multiple curvatures, capturing mixed topologies with unprecedented granularity. Parallel work on “graph foundation models” seeks a universal vocabulary that stabilizes learning across datasets \cite{foundationalgraph}. When coupled with efficient global–local backbones such as SGFormer, these advances deliver both scalability and nuance. Together they herald a future where graph learners reason quickly, respect complex structure, and open exciting avenues for real-world, billion-scale applications.

\section{Background}
In this section, we present some definitions and basic concepts on manifolds and node embeddings.
\subsection{Preliminaries on Manifolds}
Manifolds furnish a rigorous geometric setting for modern graph-representation
pipelines.  One may first
\emph{tokenise} each node $v\in V(G)$ by projecting its raw features onto an
appropriate manifold $\mathcal M$; the resulting coordinates serve as the node
embedding that is subsequently processed by the transformer layers.  Three
families of manifolds are used in this work, Stiefel, Grassmann, and
general Riemannian manifolds---each introducing a distinct geometric bias that
can improve expressiveness or training stability.

\subsection{Manifolds}
A \emph{manifold} is a topological space that is locally homeomorphic to
$\mathbb R^{n}$.  When the transition maps are smooth we obtain a \emph{smooth
manifold}; endowing it with an inner‐product field on the tangent bundle yields
a \emph{Riemannian manifold} $(\mathcal M,g)$.  The metric $g$ provides
\begin{itemize}
  \item tangent vectors $u,v\in T_{p}\mathcal M$ at every point $p\in\mathcal M$;
  \item intrinsic notions of length, angle, and volume;
  \item geodesics $\gamma$, i.e.\ curves minimising the Riemannian length functional.
\end{itemize}

These structures allow Euclidean learning algorithms to be ported to
non‐Euclidean settings: gradients are interpreted in the tangent spaces and
mapped back with a \emph{retraction} (or the exact exponential map when
available).

\subsection{Riemannian Manifolds}
A \emph{Riemannian manifold} is a pair $(\mathcal M,g)$ where
\[
  g_{p}\;:\;
  T_{p}\mathcal M \times T_{p}\mathcal M
  \;\longrightarrow\;
  \mathbb R,
  \qquad
  g_{p}(u,v) = \langle u,v\rangle_{p},
\]
is a smoothly varying positive‐definite inner product.
Curved spaces such as the $n$‐dimensional hyperbolic ball $\mathbb H^{n}$ or
the positive‐definite cone $\mathcal S_{++}^{n}$ capture hierarchical or
covariance‐type relations between nodes more naturally than any flat Euclidean
embedding.

\subsection{Stiefel Manifold \texorpdfstring{$V_{k}(\mathbb R^{n})$}{Vk(Rn)}}
\[
  V_{k}(\mathbb R^{n})
  \;=\;
  \bigl\{
    X\in\mathbb R^{n\times k}\;\bigl|\;
    X^{\!\top}X = I_{k}
  \bigr\}.
\]
Elements are ordered orthonormal $k$‐frames in $\mathbb R^{n}$.
Projecting a node’s feature matrix onto $V_{k}(\mathbb R^{n})$ enforces
orthogonality among latent directions---a useful inductive bias for multi‐head
attention.  Adaptive optimisers such as \emph{Riemannian Adam}
(Sec.~\ref{subsec:riemannian-adam}) often converge faster on $V_{k}$ than
naïve Euclidean Adam plus projection.

\subsection{Grassmann Manifold \texorpdfstring{$G_{k}(\mathbb R^{n})$}{Gk(Rn)}}
\[
  G_{k}(\mathbb R^{n})
  \;=\;
  V_{k}(\mathbb R^{n}) \big/ O(k),
\]
the quotient that discards the ordering of the frame and retains only the
$k$‐dimensional subspace it spans.  A point is an equivalence class
$[X] = \{\,XQ : Q\in O(k)\,\}$.  Encoding nodes on $G_{k}(\mathbb R^{n})$
is appropriate when only the subspace---not the basis---should matter (e.g.\
sets of symmetric attributes).


\subsection{Node Embedding Scheme}
Node features in real-world graph datasets rarely encode essential structural signals—such as node degree, temporal stamps, spatial coordinates, or higher-order neighbourhood patterns. This informational gap motivates the search for principled embedding schemes that can supply the missing topology to downstream models. \emph{GraphMoRE}~\cite{graphmore} addresses this problem by explicitly modelling topological heterogeneity: a topology-aware gating mechanism analyses each node’s local geometry and dispatches it to the most suitable embedding sub-space.

To accommodate the wide variety of local curvatures observed in complex graphs, GraphMoRE deploys a collection of Riemannian “experts’’ whose mixed-curvature spaces are stitched together into a single, node-adaptive manifold. A lightweight alignment module then normalises pairwise distances across these heterogeneous patches, guaranteeing geometric consistency throughout the embedding space. Extensive experiments confirm that this design substantially improves both adaptability and predictive performance across graphs drawn from diverse application domains, establishing GraphMoRE as a compelling foundation for topology-aware graph learning.

\begin{algorithm}[tb]
   \caption{Training procedure for R-SGFormer(S) and R-SGFormer(G)}
   \label{alg:sgformersg}
\begin{algorithmic}
   \STATE {\bfseries Input:} node–feature matrix $X$ and adjacency matrix $A$.
   \STATE Split the data into $M$ mini-batches: $\{(X_1, A_1), (X_2, A_2), \dots, (X_M, A_M)\}$.
   \FOR{each mini-batch $(X_m, A_m)$}
       \STATE (1)(a) {\it Stiefel option:} form $q = W_q X_m$ and $k = W_k X_m$, apply QR decomposition for $q$ and $k$, and keep the $Q$ factors.
       \STATE (1)(b) {\it Grassmann option:} form $q = W_q X_m$ and $k = W_k X_m$, apply singular-value decomposition  for $q$ and $k$, and keep the $U$ factors.
       \STATE (2) Use Linear Attention to compute $Z_0$.
       \STATE (3) Pass $(X_m, A_m)$ through the GNN to obtain $Z$.
       \STATE (4) Combine $Z$ and $Z_0$ by weighted summation to produce the final representation $Y$.
       \STATE (5) Generate predictions $\hat{Y}_m$ from $Y$.
       \STATE (6) Add the regularisation term $\lambda\,(Y^\top Y - I)$ to the loss.
   \ENDFOR
   \STATE {\bfseries Output:} predicted values $\{\hat{Y}_m\}$ for all mini-batches.
\end{algorithmic}
\end{algorithm}

\begin{algorithm}[tb]
   \caption{Training process for R-SGFormer}
   \label{alg:R-SGFormer}
\begin{algorithmic}
   \STATE {\bfseries Input:} graph $G$; number of experts $K$; initial curvature set $C$; total epochs $T$
   \STATE {\bfseries Output:} node--label predictions obtained by SGFormer on the cross-attended combination of raw node features and GraphMoRE embeddings
   \STATE Initialise the set of Riemannian experts $E$
   \STATE Sample local topological subgraphs $S$ for all nodes
   \FOR{$t = 1, 2, \dots, T$}
       \FOR{each expert $e \in E$}
           \STATE Compute the expert output $Z_{e}$
       \ENDFOR
       \STATE Derive local topology descriptors for all nodes
       \STATE Compute the gating weights $W$ that assign experts to individual nodes
       \FOR{each node pair $(u,v)$ in $G$}
           \STATE Obtain the aligned expert weights $W_{(u,v)}$
           \STATE Evaluate the squared embedding distance $d^{2}(u,v)$
       \ENDFOR
       \STATE Update all parameters by minimising the joint loss comprising classification error, gating entropy, and expert regularisation
   \ENDFOR
\end{algorithmic}
\end{algorithm}

\begin{figure*}[ht]
\vskip 0.2in
\begin{center}
\centerline{\includegraphics[width=0.9\textwidth]{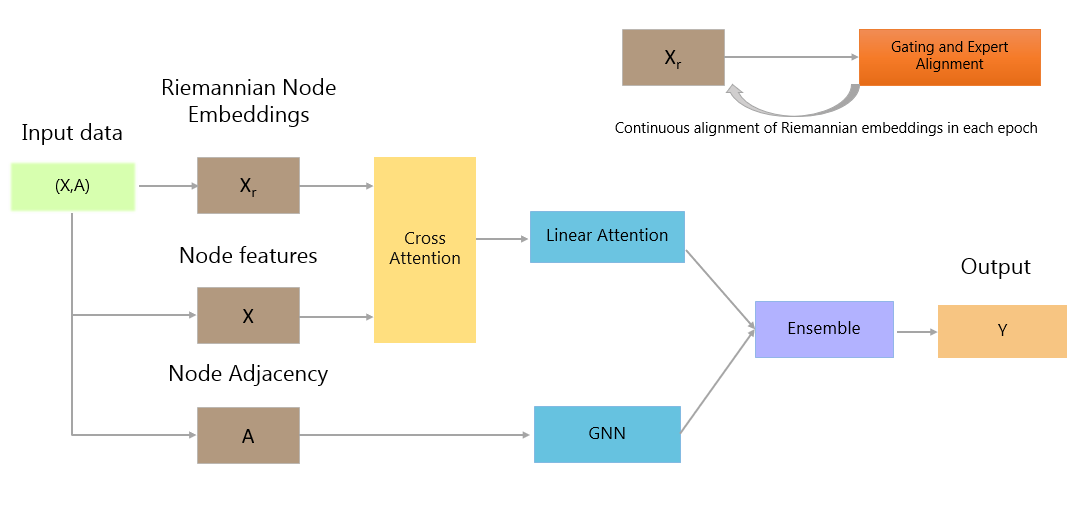}}
\caption{R-SGFormer Architecture}
\label{icml-historical}
\end{center}
\vskip -0.2in
\end{figure*}

\section{R-SGFormer}
Transformer-style architectures face a well-known scalability barrier in large-scale graph representation learning: the self-attention operator incurs \(\mathcal{O}(N^{2})\) time and memory, which rapidly becomes prohibitive for graphs containing millions or billions of nodes.  \textbf{SGFormer}~\cite{sgformer} alleviates this bottleneck by interleaving a single-layer \emph{linear} attention module with a lightweight Graph Neural Network (GNN).  The resulting hybrid preserves much of the expressive power of self-attention while reducing computational overhead to \(\mathcal{O}(N)\).  Exhaustive tests on standard node-property-prediction benchmarks demonstrate that SGFormer matches—or surpasses—the accuracy of full Transformer baselines on medium-sized graphs and remains tractable on graphs with up to \(10^{8}\) nodes, yielding order-of-magnitude speed-ups in both training and inference.

Despite these advantages, several avenues remain open for further improvement.\\
\textbf{Topology-aware queries and keys.}  
  The attention branch currently consumes only raw node features, ignoring edge attributes and higher-order structural descriptors.  Injecting curvature-adaptive or positional embeddings into the query–key pairs could furnish richer contextual signals for the attention scores.\\
\textbf{Geometric regularisation of latent space.}  
  The model does not leverage manifold regularisation or other geometric priors capable of imposing global structural coherence on the hidden representations.  Incorporating Riemannian embedding strategies, curvature-mixing experts, or spectral regularisers may enhance both expressiveness and generalisation, particularly in graphs exhibiting pronounced local heterogeneity.

\subsection{Stiefel and Grassmann Manifold Projection and Regularization}

Stiefel manifolds provide a principled means of enforcing orthogonality in both the input and latent spaces of graph models.  Empirical studies reveal that node–feature matrices in standard benchmarks naturally cluster near orthogonal subspaces: replacing the raw feature matrix \(X\) by the orthogonal factor \(Q\) from its QR decomposition leaves SGFormer’s predictive accuracy virtually unchanged.  This invariance indicates that the data’s intrinsic structure is already aligned with orthogonal directions, motivating explicit projection onto the Stiefel manifold together with an accompanying regulariser.  When such a regularisation term is applied to the final ensemble representation, the output is nudged toward an orthogonal matrix, encouraging the attention and GNN branches to supply complementary, nearly orthogonal feature sets.  A parallel argument holds for Grassmann manifolds: projecting inputs onto a Grassmannian subspace—and softly constraining the outputs to remain there—preserves global structural information that might otherwise be lost.

Building on these observations, we augment the base SGFormer with manifold-aware components.  The variant that employs Stiefel projections and regularisation is denoted \textbf{R-SGFormer(S)}, while the Grassmann-based counterpart is referred to as \textbf{R-SGFormer(G)}. Detailed algorithm is provided(as shown in Algorithm \ref{alg:sgformersg}) and experiments are also performed with this technique.

\subsection{Riemannian Node Embeddings}
Empirical gains obtained from Stiefel- and Grassmann-based projections suggest that curvature information should be injected directly at the node level.  Consequently, we adopt \emph{GraphMoRE} as a Riemannian‐embedding front end for SGFormer and refer to the resulting model as \textbf{R-SGFormer}.  GraphMoRE supplies a mixed‐curvature embedding for each node, capturing fine‐grained geometric cues that the original features alone cannot express.  In preliminary experiments, substituting these embeddings for—or concatenating them with—one‐hot encodings yields a consistent performance uplift over the vanilla SGFormer backbone.

R-SGFormer processes the two feature streams through a cross‐attention block that fuses the raw attributes with their Riemannian counterparts before passing the combined representation to the linear‐attention layer of SGFormer (Algorithm~\ref{alg:R-SGFormer}).  This architecture enables the model to learn \emph{when} and \emph{how} to exploit curvature‐aware information while still benefiting from the scalability of linear attention.  Our emprical experiments study shows that preserving manifold structure at both the input and output stages is crucial: input‐level preservation supplies a well‐structured signal to the network, whereas output‐level regularisation encourages the ensemble representation to remain in a geometrically coherent subspace, jointly boosting generalisation.

\begin{table*}
\caption{Performance comparison of node classification for various methods on multiple datasets. Bold values represent the best performance for each metric.}
\begin{center}
\begin{small}
\begin{sc}
\scalebox{0.85}{
\begin{tabular}{lccccccccccc}
\toprule
Method & \multicolumn{2}{c}{Cora} & \multicolumn{2}{c}{Citeseer} & \multicolumn{2}{c}{Airport} & \multicolumn{2}{c}{PubMed}\\
\cmidrule(lr){2-3} \cmidrule(lr){4-5} \cmidrule(lr){6-7} \cmidrule(lr){8-9}
 & W-F1 & M-F1 & W-F1 & M-F1 & W-F1 & M-F1 & W-F1 & M-F1\\
\midrule
GCN & 79.41$\pm$1.25 & 78.92$\pm$1.04 & 65.92$\pm$2.13 & 62.46$\pm$1.73 & 81.38$\pm$0.82 & 77.44$\pm$0.82 & 75.95$\pm$0.40 & 75.74$\pm$0.34\\
GAT & 77.49$\pm$1.11 & 76.91$\pm$1.04 & 65.43$\pm$1.46 & 61.87$\pm$1.38 & 83.07$\pm$2.19 & 81.27$\pm$1.27 & 75.33$\pm$0.89 & 75.08$\pm$0.84\\
\midrule
HGCN & 77.78$\pm$0.63 & 73.94$\pm$0.61 & 67.51$\pm$0.81 & 66.50$\pm$0.83 & 88.82$\pm$1.46 & 81.23$\pm$0.26 & 78.69$\pm$0.59 & 77.78$\pm$0.50\\
$\kappa$-GCN & 78.67$\pm$1.00 & 78.01$\pm$0.73 & 63.88$\pm$0.69 & 60.28$\pm$0.60 & 87.40$\pm$1.64 & 84.33$\pm$2.08 & 77.31$\pm$1.71 & 76.37$\pm$1.54\\
Q-GCN & 75.97$\pm$0.97 & 74.03$\pm$0.96 & 68.89$\pm$1.70 & 62.68$\pm$1.48 & 89.66$\pm$0.67 & 85.46$\pm$0.86 & 80.61$\pm$0.81 & 79.55$\pm$0.91\\
MotifRGC & 80.91$\pm$0.64 & 80.19$\pm$0.63 & 68.31$\pm$1.41 & 63.11$\pm$1.41 & 79.30$\pm$6.23 & 79.30$\pm$6.23 & 79.13$\pm$1.37 & 77.71$\pm$1.34\\
\midrule
\tiny{GraphMoRE$_\text{GCN}$} & 80.51$\pm$1.04 & 79.44$\pm$0.77 & \textbf{69.73$\pm$0.70} & \textbf{65.98$\pm$0.67} & 91.75$\pm$0.93 & 90.19$\pm$0.86 & 76.50$\pm$0.76 & 75.60$\pm$0.76\\
\tiny{GraphMoRE$_\text{GAT}$} & 79.81$\pm$0.71 & 78.53$\pm$0.85 & 68.59$\pm$1.64 & 64.70$\pm$1.74 & 91.06$\pm$1.52 & 90.50$\pm$1.78 & 76.46$\pm$0.76 & 75.46$\pm$0.76\\
\tiny{GraphMoRE$_\text{SAGE}$} & 81.42$\pm$0.6 & 80.32$\pm$0.56 & 69.40$\pm$0.82 & 65.71$\pm$1.12 & 92.32$\pm$0.70 & 91.33$\pm$0.46 & 77.30$\pm$0.81 & 76.53$\pm$0.73\\
\midrule
\tiny{\textbf{R-SGFormer$_{(Ours)}$}} & \textbf{82.44$\pm$0.68} & \textbf{80.66$\pm$0.63} & 66.26$\pm$0.34 & 64.46$\pm$0.47 & \textbf{93.53$\pm$0.51} & \textbf{93.18$\pm$0.21} & \textbf{81.01$\pm$0.21} & \textbf{80.21$\pm$0.73}\\
\bottomrule
\end{tabular}
}
\end{sc}
\end{small}
\end{center}
\label{tab:results}
\end{table*}

\subsection{Algorithms}
Our empirical study evaluates three SGFormer–based variants.  
\textbf{R-SGFormer(S)} augments SGFormer with a Stiefel-manifold projection of the query–key matrices and an orthogonality regulariser on the final representation.  
\textbf{R-SGFormer(G)} applies the same scheme using a Grassmann-manifold projection.  
\textbf{R-SGFormer} (GraphMoRE\,{+}\,SGFormer) replaces fixed projections with curvature-adaptive node embeddings generated by GraphMoRE.

Algorithm~\ref{alg:sgformersg} outlines the training loop for \textbf{R-SGFormer(S)} and \textbf{R-SGFormer(G)}.  
For each mini-batch, queries and keys are orthogonalised via QR (Stiefel) or thin SVD (Grassmann), yielding structured factors for the Linear-Attention module.  
The resulting auxiliary representation $Z_0$ is passed, together with $(X_m,A_m)$, into the GNN to produce $Z$.  
The two streams are fused by summation to obtain the final embedding $Y$.  
A task-specific loss is augmented with an orthogonality penalty $\lambda\lVert Y^{\top}Y-I\rVert_F^{2}$, encouraging the model to learn complementary, near-orthogonal features.

Algorithm~\ref{alg:R-SGFormer} summarises the complete \textbf{R-SGFormer} pipeline.  
Training begins by initialising $K$ Riemannian experts, each with a curvature selected from the set $C$, and by sampling local subgraphs for every node.  
At every epoch the experts generate embeddings $Z_e$, local topology descriptors are computed, and a gating network assigns expert weights to individual nodes.  
Aligned weights and pair-wise embedding distances are evaluated for each node pair $(u,v)$, after which SGFormer performs cross-attention between the raw node features and the expert embeddings.  
Model parameters are updated by minimising a composite objective that combines classification error, gating entropy, and expert regularisation, yielding geometry-aware node representations that adapt to local graph heterogeneity.

\begin{table*}
\caption{Node classification accuracies for SGFormer, R-SGFormer(S) and R-SGFormer(G)}
\label{sample-table}
\begin{center}
\begin{small}
\begin{sc}
\begin{tabular}{lccccccc}
\toprule
Dataset & Cora & CiteSeer & PubMed & Actor & Squirrel & Chameleon & Deezer \\
\midrule
\# nodes & 2,708 & 3,327 & 19,717 & 7,600 & 2,223 & 890 & 28,281 \\
\# edges & 5,278 & 4,552 & 44,324 & 29,926 & 46,998 & 8,854 & 92,752 \\
\midrule
GCN & 81.6 $\pm$ 0.4 & 71.2 $\pm$ 1.0 & 78.8 $\pm$ 0.6 & 30.1 $\pm$ 0.2 & 38.6 $\pm$ 1.8 & 41.3 $\pm$ 3.0 & 62.7 $\pm$ 0.7 \\
GAT & 80.1 $\pm$ 0.6 & 70.9 $\pm$ 1.1 & 78.7 $\pm$ 0.6 & 27.0 $\pm$ 0.9 & 31.6 $\pm$ 2.3 & 39.0 $\pm$ 3.3 & 61.0 $\pm$ 0.9 \\
SGC & 80.1 $\pm$ 0.6 & 71.9 $\pm$ 0.7 & 78.7 $\pm$ 0.6 & 30.8 $\pm$ 1.0 & 39.6 $\pm$ 3.3 & 39.0 $\pm$ 3.3 & 61.0 $\pm$ 0.9 \\
JKNet & 81.3 $\pm$ 0.7 & 71.9 $\pm$ 0.7 & 78.1 $\pm$ 0.6 & 30.5 $\pm$ 1.2 & 39.3 $\pm$ 1.6 & 41.3 $\pm$ 3.3 & 62.6 $\pm$ 0.8 \\
APPNP & 83.3 $\pm$ 0.5 & 71.8 $\pm$ 0.5 & 80.1 $\pm$ 0.6 & 29.4 $\pm$ 0.3 & 30.2 $\pm$ 1.4 & 40.2 $\pm$ 3.9 & 64.5 $\pm$ 0.8 \\
H$_2$GCN & 82.1 $\pm$ 0.3 & 72.4 $\pm$ 0.7 & 79.5 $\pm$ 0.5 & 30.8 $\pm$ 1.2 & 38.4 $\pm$ 1.5 & 41.2 $\pm$ 3.8 & 62.6 $\pm$ 0.7 \\
SIGN & 82.1 $\pm$ 0.3 & 72.4 $\pm$ 0.7 & 79.5 $\pm$ 0.5 & 30.8 $\pm$ 1.2 & 38.4 $\pm$ 1.5 & 41.2 $\pm$ 3.8 & 62.6 $\pm$ 0.7 \\
CPGNN & 81.9 $\pm$ 0.4 & 72.1 $\pm$ 0.6 & 79.0 $\pm$ 0.5 & 30.6 $\pm$ 1.1 & 38.7 $\pm$ 1.4 & 41.3 $\pm$ 3.6 & 62.4 $\pm$ 0.7 \\
GloGNN & 81.9 $\pm$ 0.4 & 72.1 $\pm$ 0.6 & 79.0 $\pm$ 0.5 & 36.4 $\pm$ 1.6 & 35.7 $\pm$ 1.8 & 40.2 $\pm$ 3.9 & 65.8 $\pm$ 0.8 \\
\midrule
NodeFormer & 82.2 $\pm$ 0.8 & 72.5 $\pm$ 1.1 & 79.9 $\pm$ 1.0 & 35.9 $\pm$ 1.1 & 41.8 $\pm$ 2.2 & 44.9 $\pm$ 3.9 & 66.4 $\pm$ 0.7 \\
\midrule
SGFormer & 84.5 $\pm$ 0.8 & 72.6 $\pm$ 0.2 & 80.3 $\pm$ 0.6 & 37.9 $\pm$ 1.1 & 41.8 $\pm$ 2.2 & 44.9 $\pm$ 3.9 & 67.1 $\pm$ 1.1 \\
\midrule
R-SGFormer(S) & 86.2 $\pm$ 0.8 & 73.2 $\pm$ 0.7 & 81.5 $\pm$ 1.9 & 40.1 $\pm$ 1.5 & 42.1 $\pm$ 2.5 & 45.7 $\pm$ 1.4 & 68.9 $\pm$ 0.4 \\
\midrule
R-SGFormer(G) & 85.2$\pm$ 0.2 & 74.7$\pm$ 1.7 & 80.6$\pm$ 1.3 & 37.8$\pm$ 1.3 & 42.3$\pm$ 3.2 &
41.4$\pm$ 1.2 & 67.2$\pm$ 2.4\\
\bottomrule
\end{tabular}
\end{sc}
\end{small}
\end{center}
\end{table*}

\section{Experiments}
All experiments were conducted on a single NVIDIA A100 GPU (80 GB HBM) using both homophilic and heterophilic node-classification benchmarks.  To facilitate a fair comparison, we report the best numbers available for strong, balanced baselines drawn from the SGFormer and GraphMoRE literature; models that excel only on either homophilic \emph{or} heterophilic graphs were excluded.  Homophilic graphs contain edges that preferentially link nodes with similar attributes (e.g.\ social networks), whereas heterophilic graphs exhibit the opposite pattern (e.g.\ user–item bipartite graphs in recommender systems).

\vspace{0.3em}
\noindent\textbf{Hyper-parameter search.}
Curvature candidates for GraphMoRE experts were fixed to ${-3,-1,0,1,3}$.  For SGFormer we swept the learning rate over ${0.1,,0.01,0.001}$ and the graph-mixing weight over ${0.1,\dots,0.9}$.  The orthogonality coefficient $lambda$ in R-SGFormer(S) and R-SGFormer(G) was selected from ${0.1,,0.01,,0.0001}$.  All models were optimised with ADAM \cite{adam}; GraphMoRE’s internal experts used the Riemannian-Adam variant of \cite{riemannianadam}.  Training epochs ranged from 50 to 500, with longer schedules required when expert embeddings caused oscillatory validation curves.

\vspace{0.3em}
\noindent\textbf{Stiefel vs.\ Grassmann projections.}
R-SGFormer(S) attains the highest accuracies on \textsc{Cora} (86.2\%), \textsc{CiteSeer} (73.2\%), \textsc{PubMed} (81.5\%), \textsc{Actor} (39.4\%), \textsc{Squirrel} (42.3\%), and \textsc{Deezer} (68.9\%).  R-SGFormer(G) is superior on \textsc{CiteSeer} (74.7\%), \textsc{Chameleon} (41.4\%), and approaches parity elsewhere.  The Stiefel variant is favoured when a graph’s latent structure is naturally close to an orthogonal frame; the Grassmann variant excels when rank-invariant features dominate, as its projection preserves subspace information rather than basis orientation.

\vspace{0.3em}
\noindent\textbf{R-SGFormer (GraphMoRE + SGFormer).}
The combined model consistently surpasses both SGFormer and the original GraphMoRE adapters for GCN, GAT, and SAGE.  In terms of weighted / macro F1, R-SGFormer records 82.44 / 80.66\% on \textsc{Cora}, 93.53 / 93.18\% on \textsc{Airport}, and 81.01 / 80.21\% on \textsc{PubMed}.  Performance on \textsc{CiteSeer} is lower (66.26 / 64.46\%), likely due to that dataset’s extreme sparsity, yet the model remains competitive.

\vspace{0.3em}
\noindent\textbf{Insights.}
Across every dataset, both R-SGFormer(S) and R-SGFormer(G) improve on the plain SGFormer backbone, demonstrating the benefit of manifold-aware projections. R-SGFormer helps achieve balanced gains. The ensemble architecture also preserves Euclidean information through the GNN branch while injecting non-Euclidean structure via attention, validating the premise that granular Riemannian embeddings can materially strengthen transformer-style graph models.


\section{Conclusion}
It has been demonstrated that combining the granular, curvature-adaptive embeddings of \textbf{GraphMoRE} with the scalable backbone of \textbf{SGFormer} yields a principled architecture—\textbf{R-SGFormer}—that achieves or surpasses state-of-the-art accuracy on diverse node-classification benchmarks.  Beyond the full integration, our Stiefel and Grassmann variants reveal that manifold projection coupled with an orthogonality regulariser provides a lightweight yet effective route to inject geometric bias, improving generalisation while preserving SGFormer’s linear attention speed.  Collectively, these results establish a coherent framework for embedding manifold structure directly into transformer-style graph models.

A central insight is that \emph{structure preservation at both ends of the pipeline matters}: projecting inputs onto a suitable manifold and nudging gradients to remain in the same geometric subspace ensure that the network learns representations consistent with the graph’s intrinsic topology.  This heuristic affords practitioners a practical test—do orthogonal or subspace projections improve validation loss?—for selecting an appropriate manifold family.  Moreover, the scheme dovetails naturally with other ensemble settings, including collaborative-filtering transformers whose global attention patterns stand to benefit from curvature-aware tokenisation.

Opportunities for extension are abundant.  Hyperbolic or mixed-curvature transformers could replace the current Euclidean attention block, pushing scalability into the negative-curvature regime.  Lorentzian embeddings offer a tantalising path to encode temporal edges or relativistic distances, enriching spatiotemporal graphs with physically meaningful structure.  Finally, a fixed “foundational’’ library of Riemannian node embeddings—analogous to pretrained word vectors in NLP—would enable rapid transfer to ever larger graphs and tasks, while opening the door to deeper mathematical interpretability.  We believe that embedding geometry at the heart of graph transformers marks a promising frontier for representation learning.
\section*{Impact Statement}
The advancements in the field of graph based-learning will one of the most important when discovering medicines since molecules are represented using graphs and analysing complex social networks. This paper presents work whose goal is to advance the field of Machine Learning. There are many potential societal consequences  of our work, none which we feel must be specifically highlighted here.

\newpage
\bibliography{icml_rsgformer}
\nocite{graphMETRO}
\nocite{mixedcurvaturemultirelation}
\nocite{linearattentionorthogonalmemory}
\nocite{Klingenberg+1995}
\nocite{pmlr-v119-hassani20a}
\bibliographystyle{icml2025}


\end{document}